\documentclass[twocolumn,10pt]{asme2e}

\usepackage{graphicx}
\usepackage{mathtools}
\usepackage{amsfonts}
\usepackage{calrsfs}
\usepackage{cite}
\usepackage{subcaption}
\DeclareMathAlphabet{\pazocal}{OMS}{zplm}{m}{n}

\usepackage{xcolor}
\usepackage{hyperref}

\usepackage{float}

\usepackage{rotating}
\usepackage{multirow}
\usepackage{multicol}

\usepackage[utf8]{inputenc}

\usepackage{graphicx}
\usepackage{amsmath}

\usepackage[version=4]{mhchem}
\usepackage{siunitx}
\usepackage{longtable,tabularx}
\usepackage{bm}
\usepackage[inline]{enumitem}
\usepackage{array}

\confshortname{IDETC/CIE2022}
\conffullname{the ASME 2022 \\
International Design Engineering Technical Conferences \\
              and Computers and Information in Engineering Conferences}
\papernum{DRAFT DETC2022-xxxxx}
\confdate{14-17}
\confmonth{August}
\confyear{2022}
\confcity{St. Louis, MO}
\confcountry{USA}

\papernum{DETC2022-xxxxx}

\title{Design Target Achievement Index: A Differentiable Metric to Enhance Deep Generative Models in Multi-Objective Inverse Design}

\author{Lyle Regenwetter
    \affiliation{
	Dept. of Mechanical Engineering\\
	Massachusetts Institute of Technology\\
	Cambridge, MA 02139\\
    Email: regenwet@mit.edu
    }	
}
\author{Faez Ahmed
    \affiliation{
	Dept. of Mechanical Engineering\\
	Massachusetts Institute of Technology\\
	Cambridge, MA 02139\\
    Email: faez@mit.edu
    }	
}

\newcommand{\etal}{{\em et~al.}}

\newcommand{\etc}{{\em etc.}}
\newcommand{\RNum}[1]{\uppercase\expandafter{\romannumeral #1\relax}}

\usepackage[linesnumbered,ruled,vlined]{algorithm2e} 

\begin{document}

\maketitle    

%%%%%%%%%%%%%%%%%%%%%%%%%%%%%%%%%%%%%%%%%%%%%%%%%%%%%%%%%%%%%%%%%%%%%%
\begin{abstract}
{\it 
Deep Generative Machine Learning Models have been growing in popularity across the design community thanks to their ability to learn and mimic complex data distributions. While early works are promising, further advancement will depend on addressing several critical considerations such as design quality, feasibility, novelty, and targeted inverse design. We propose the Design Target Achievement Index (DTAI), a differentiable, tunable metric that scores a design's ability to achieve designer-specified minimum performance targets. We demonstrate that DTAI can drastically improve the performance of generated designs when directly used as a training loss in Deep Generative Models. We apply the DTAI loss to a Performance-Augmented Diverse GAN (PaDGAN) and demonstrate superior generative performance compared to a set of baseline Deep Generative Models including a Multi-Objective PaDGAN and specialized tabular generation algorithms like the Conditional Tabular GAN (CTGAN). We further enhance PaDGAN with an auxiliary feasibility classifier to encourage feasible designs. To evaluate methods, we propose a comprehensive set of evaluation metrics for generative methods that focus on feasibility, diversity, and satisfaction of design performance targets. Methods are tested on a challenging benchmarking problem: the FRAMED bicycle frame design dataset featuring mixed-datatype parametric data, heavily skewed and multimodal distributions, and ten competing performance objectives.

% Additionally, research in the field needs to be more rigorous and organized. Researchers developing Deep Generative Models should benchmark methods on datasets which are sufficiently challenging to ensure said method is sufficiently robust to transfer to other problems. Furthermore, researchers should adopt a standardized set of evaluation metrics to allow for fair comparisons between methods. 
}
\end{abstract}

\setlength{\belowdisplayskip}{5pt} \setlength{\belowdisplayshortskip}{5pt}
\setlength{\abovedisplayskip}{5pt} \setlength{\abovedisplayshortskip}{5pt}

%%%%%%%%%%%%%%%%%%%%%%%%%%%%%%%%%%%%%%%%%%%%%%%%%%%%%%%%%%%%%%%%%%%%%%

\section{Introduction}
Automatically creating innovative designs that outperform all existing solutions and meet complex real-world engineering constraints is the holy grail of data-driven engineering design. This is an incredibly demanding task and current design automation tools remain insufficient for full autonomy in product design. Recently, Deep Generative Models have emerged as a viable means to bring us toward this overarching design automation goal. Deep Generative Models (DGMs) refer to machine learning algorithms that leverage sequential layers to learn progressively deeper understandings of design representations. Typically, these models are trained to understand the distribution of existing designs in some design space (usually using a dataset of existing designs), then generate new designs by sampling from this learned distribution. 

DGMs are typically trained to maximize statistical similarity between distributions of generated samples and the underlying data distribution. In engineering design, design objectives and constraints make statistical similarity metrics insufficient and sometimes inappropriate. Despite this, an overwhelming majority of research in engineering design continues to optimize and evaluate methods using statistical similarity. We believe the continuation of this practice is rooted in two central challenges. Firstly, appropriate metrics to evaluate DGMs on engineering objectives such as design performance, feasibility, and novelty are poorly established. Secondly, researchers lack effective methods to build these auxiliary objectives into training procedures and instead fall back upon the established structural similarity as the central training mechanism. Addressing these two challenges is the central thrust of this paper. 

\paragraph{Contributions:}
Our key contributions are summarized as follows:
\begin{enumerate}
    \item We propose a multifaceted set of evaluation metrics for Deep Generative Models in engineering design consisting of our novel DTAI metric, as well as seven other metrics focusing on design and performance space diversity, novelty, feasibility, and target satisfaction.
    \item We introduce the Design Target Achievement Index (DTAI), a differentiable scoring metric which allows Deep Generative Models to prioritize, meet, and exceed multi-objective performance targets specified by a designer. 
    \item We augment a state-of-the-art Performance-Augmented Diverse GAN with a loss based our DTAI function and feasibility estimator and demonstrate that this framework yields significant performance improvements, such as increasing the average proportion of design targets met by 45\% and proportion of feasible designs by 30\% versus state-of-the-art tabular generation methods. To our knowledge, this proposed framework marks the first Deep Generative Method that actively optimizes for overall design performance, diversity, feasibility, and target satisfaction simultaneously. 
    \item We evaluate several existing Deep Generative Models using our proposed evaluation metrics on a challenging real-world design dataset with ten competing objectives and complicated regions of infeasibility. We demonstrate that our DTAI-augmented method significantly outperforms baseline DGMs in numerous performance metrics. 
\end{enumerate}
In the following sections, we discuss the dataset used, the methods tested, the evaluation metrics proposed, and the results of our analysis. 

\section{Review of Deep Generative Models in Engineering Design}
In a recent review of Deep Generative Models (DGMs), Regenwetter~\etal~\cite{regenwetter2021deep} discuss the application of DGMs across engineering design fields and analyze key limitations in the current state-of-the-art in DGM methodology. The authors suggest that successfully addressing several key challenges will be essential in the continued development of DGMs for engineering design. Four of these challenges are design quality, design novelty, more robust design representation methods, and targeted inverse design. In this section, we briefly summarize the state of the current research, as well as key drivers behind each of these four challenges. For a more detailed review and discussion, we refer the reader to ~\cite{regenwetter2021deep}. 

\paragraph{Design Quality:}\label{Quality} Design 1uality (Performance) is an essential component of the design process. Design quality may be comprised of many diverse (and sometimes competing) objectives. For example, the key measures of a bicycle's quality may include weight, cost efficiency, structural integrity, aesthetics, aerodynamics, and ergonomics. Designs must be high-performing to be competitive, economically viable, and positively impact customers. As Regenwetter~\etal~\cite{regenwetter2021deep} note, most existing Deep Generative Models fail to account for design performance of any kind, perhaps since most DGMs are adapted from other disciplines such as computer vision, where notions of performance are limited to reconstruction quality or ``realism,'' of generated samples. In general, approaches to incorporate design quality into DGMs fall into three broad categories: 
\begin{enumerate}
    \item Building performance-estimating objectives into the training function of the DGM~\cite{chen2021padgan, chen2021mopadgan, chen2021geometry, ZHANG2021103041} allows the DGM to directly optimize for performance during training. For example, Ahmed \& Chen ~\cite{chen2021padgan, chen2021mopadgan}, train a surrogate model which estimates aerodynamic lift and drag and build this surrogate into the overall loss calculation in order to generate airfoil designs with high lift/drag ratios.
    \item Iteratively training the DGM on datasets augmented with high-perfoming generated designs~\cite{yoo2021integrating, oh2019deep, shu20203d, Fujita2021Design, dering2018physics} can bias generative models toward higher-performing regions of the design space. For example, Shu~\etal~\cite{shu20203d} generate 3D models of aircrafts, computationally evaluate their aerodynamic properties, then add high performing models to their dataset before retraining their model.
    \item Fitting surrogate models to link a learned latent design representation with performance parameters~\cite{rawat2019application, guo2018indirect, dong2019inverse, li2020designing, liu2020hybrid} can allow direct gradient-based optimization of latent vectors to decode into designs. For example, Li~\etal~\cite{li2020designing} use a variational autoencoder to learn a latent embedding of phononic crystal designs, then optimize latent variables using a Deep Neural Network by mapping latent vectors to target band gap values.
\end{enumerate} 
Our proposed Design Target Achievement Index (DTAI) builds on the first approach by directly aggregating performance estimates into the novel DTAI training loss. 

\paragraph{Design Novelty:} Novelty is another essential objective in the design process. For designs to improve upon existing products, they must introduce some novelty as compared to existing designs. Novelty is also an essential prerequisite for designers to market and sell products that do not infringe on the intellectual property of other designers. An inherent challenge with the construction of most DGM frameworks lies in their central objective of mimicking an existing dataset of designs. Once again, this trend traces back to the origins of DGMs in computer vision, where mimicking existing images is perfectly desirable. In design, we may prefer to distance generated designs from existing ones while still being realistic. Several works have partially addressed the challenge of novelty by incorporating humans into the generative design process~\cite{lee2019case, wang_peng_li_chen_wu_wang_childs_guo_2019}. However, relying on humans for novelty severely detracts from the autonomy of DGMs, which is arguably their key strength. A very limited number of DGM methodologies have been proposed to directly encourage the novelty of designs autonomously. Chen \& Ahmed propose a framework called the Performance-Augmented Diverse GAN (PaDGAN)~\cite{chen2021padgan} which simultaneously optimizes for performance and novelty by injecting these objectives into the training process. They extend this work to a multi-objective setting in~\cite{chen2021mopadgan}, but only consider a relatively limited test case with two objectives. 

\paragraph{Design Representation:} Designs can be digitized using countless representation schemes, and as Regenwetter~\etal~\cite{regenwetter2021deep} note, many commonly used parameterizations severely limit the usability of generated designs in downstream tasks. Representations like images and voxelizations are easy to train on due to their spatially-structured properties, but generated designs are difficult to evaluate using computational analysis tools like Finite Element Analysis, and even more challenging to physically fabricate. DGMs trained on design representations using interpretable physical dimensions of products are much more viable in downstream tasks, but are significantly more challenging to train, due to the lack of structure in their representation, mixed datatypes, and heavily unpredictable distributions~\cite{BIKED}. These challenges of training DGMs on tabular parametric data are noted across domains~\cite{xu2019modeling}.

\paragraph{Inverse Design:}
Most existing DGMs in engineering design lack any mechanism to condition design generation toward a specific set of designer-specified performance targets. The recently proposed Performance-conditioned Diverse GAN (PcDGAN)~\cite{heyrani2021pcdgan} generates designs whose performance \textit{exactly} matches a \textit{single} designer-specified target performance. In this work, however, we consider design problems in which the algorithm attempts to \textit{exceed} a set of \textit{multiple} minimum performance targets, a task which Deep Generative Models have, to our knowledge, not yet addressed. However, many approaches from the multi-objective optimization field consider such minimum performance targets, albeit with some limitations. A common approach sees targets handled as hard constraints which take precedence over any sort of design optimality. This approach is seen in well-known optimization algorithms like NSGA-II\cite{deb2002fast}, which prioritize the resolution of constraint violation before moving on to optimization. While this rigid treatment of design targets can be helpful, hard constraints lack the nuance afforded by softer design objectives, a challenge which we address with our proposed Design Target Achievement Index.

\section{Evaluating Generative Methods}
Establishing metrics to evaluate DGMs on training objectives beyond structural similarity is essential. In engineering design, we are often given performance targets during design tasks that constitute the minimum performance necessary to meet design goals. In practice, designers implicitly adapt their design process based on these minimum performance targets in ways that are difficult to quantify:
\begin{enumerate}
    \item If a design is underperforming the performance target in a particular design objective, design iterations should focus on improving performance in this objective
    \item If a design is drastically outperforming the performance target in a particular design objective, design iterations should not prioritize the further improvement of this objective
    \item Design metrics are typically weighted adaptively to match the relative importance of different design targets. 
\end{enumerate}
While we can semantically describe these phenomena, the existing tools used in multi-objective design optimization like hypervolume fail to capture their nuance. Instead, we propose the Design Target Achievement Index, a fast, differentiable scoring method of design performance that addresses each of these concerns by adaptively weighting objectives and specifically rewarding designs that satisfy design targets (Sec.~\ref{sec:aggregate}). We then discuss several other metrics measuring various other aspects of design generation, such as diversity and feasibility. 

\subsection{Design Target Achievement Index (DTAI)} \label{sec:aggregate}
\begin{figure}
    \centering
    \includegraphics[width=0.5\textwidth]{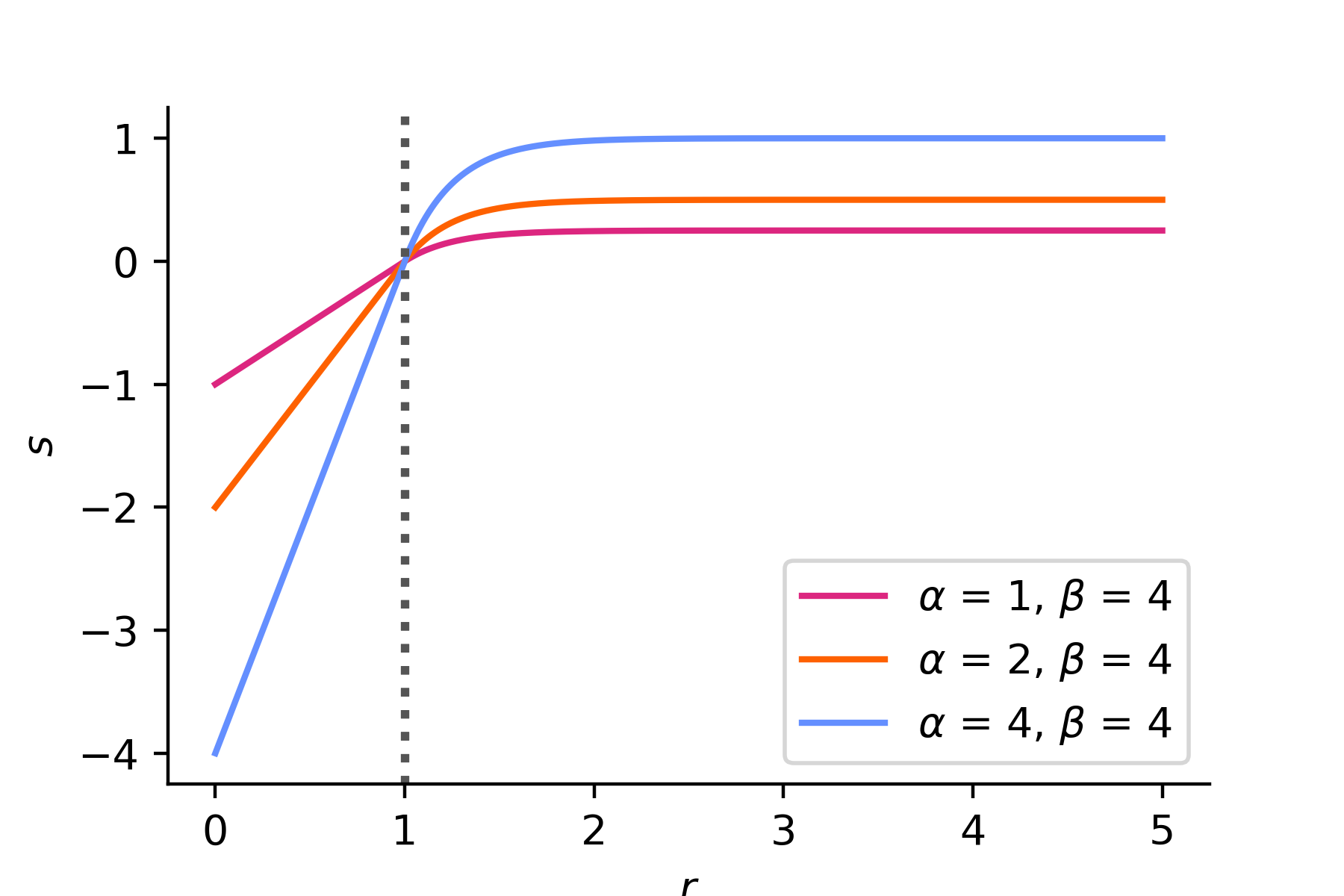}
    \caption{Effect of $\alpha$ parameter on Design Target Achievement Index.}
    \label{fig:alphamod}
\end{figure}

\begin{figure}
    \centering
    \includegraphics[width=0.5\textwidth]{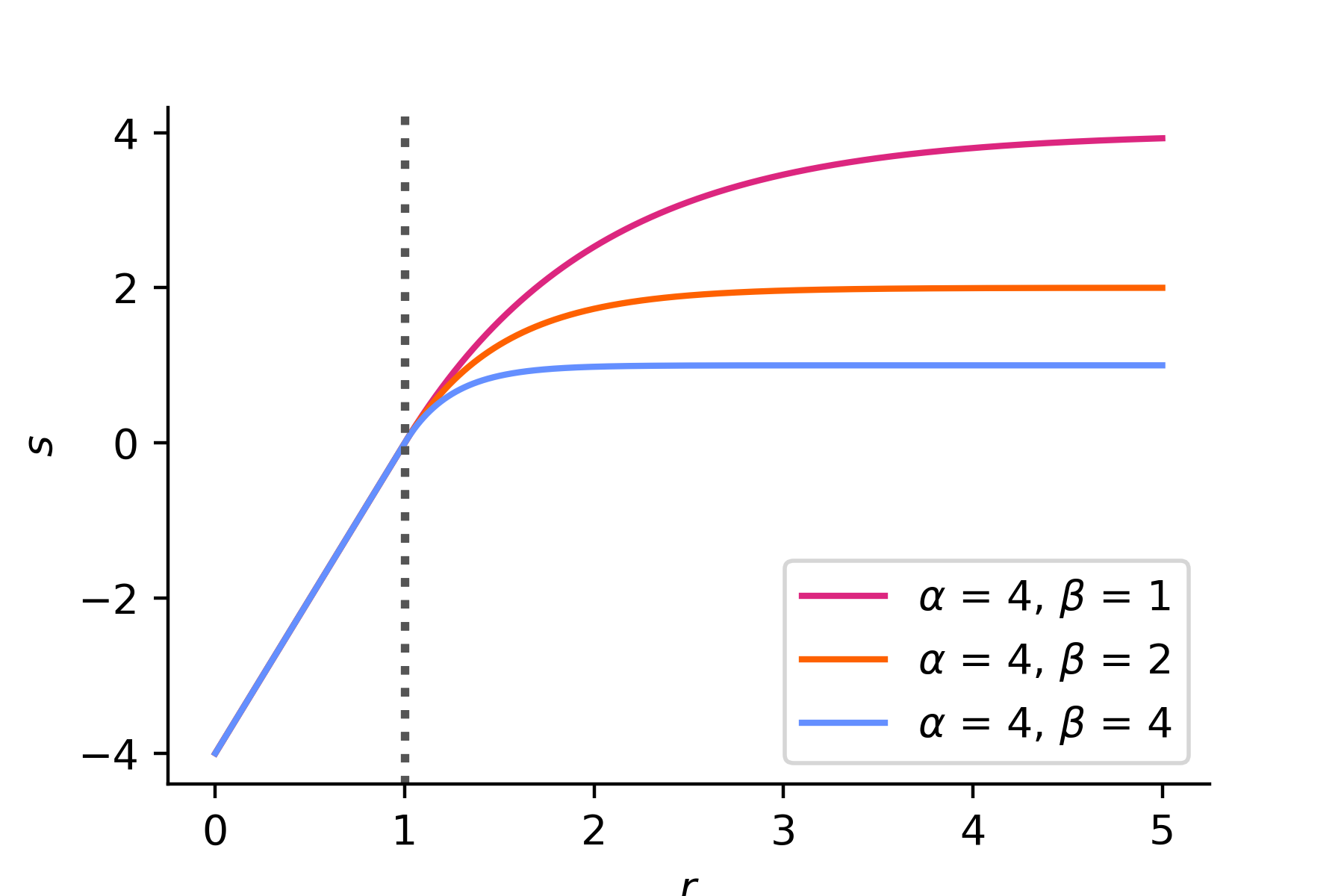}
    \caption{Effect of $\beta$ parameter on Design Target Achievement Index.}
    \label{fig:betamod}
\end{figure}

We propose a novel approach to quantify a design's performance with respect to multiple performance targets in a single metric. Consider a design, $i$, and let its performance be $p_{i,k}$ with respect to a particular performance target, $t_k$, for $k\in\{1 ... T\}$ where $T$ is the number of design objectives (i.e. weight, safety factor,~\etc). In any given objective, $k$, we desire our design's performance to exceed the performance target: $p_{i,k}\geq t_{k}$. In this formulation, we require that design performance and targets be strictly positive and the objectives maximized, however, most design objectives can be trivially reformulated as such. We express the design's performance with respect to each target as the ratio, $r_{i,k}$, between performance and the performance target:
\begin{equation}
    r_{i,k}=\frac{p_{i,k}}{t_{k}}
\end{equation}
When our design's performance exceeds the target, $r_{i,k}\geq1$. We propose the following piecewise scoring function to compute an individual target achievement score, $s_{i,k}$ in terms of $r_{i,k}$.
\begin{equation}
    s_{i,k}=\begin{cases} 
          \alpha_k(1-r_{i,k}) & r_{i,k}\leq 1 \\
          \frac{\alpha_k}{\beta_k}(1-e^{\beta_k(1-r_{i,k})}) & r_{i,k}> 1 
    \end{cases}
\end{equation}
This function is parameterized by two tuning factors, $\alpha_k$ and $\beta_k$, which are best visualized graphically. $\alpha_k$ adjusts the importance of target $k$ and can be thought of as the key weighting factor used to tune the relative importance between performance targets. $\beta_k$ is slightly more subtle and reflects the relative importance of further improvement to the objective after the performance target is met. Continuing to optimize some objectives may be more helpful than others and this parameter gives designers the ability to adjust for this nuance. For most design problems, we recommend $\beta_k$ values around $3-5$. Figures~\ref{fig:alphamod} and~\ref{fig:betamod} show the effect of adjusting $\alpha_k$ and $\beta_k$. 


The individual scores $s_{i,k}$ for each objective are easily summed into an aggregate score, then scaled by the theoretical minimum and maximum values for the sum, $s_{min}$ and $s_{max}$. This final result is our proposed Design Target Achievement Index, $s_{DTAI}$.
\begin{equation}
    s_{DTAI, i}=\frac{(\sum_{k=1}^n s_{i,k}) -s_{min}}{s_{max}-s_{min}}
\end{equation}
$s_{min}$ and $s_{max}$ are easily derived as:
\begin{equation}
    s_{min}=-\sum_{k=1}^T \alpha_k, \,\,\, s_{max}=\sum_{k=1}^T \frac{\alpha_k}{\beta_k}
\end{equation}
The proposed scoring function has numerous benefits that make it desirable both in an evaluation setting and as an objective function for the training of Deep Generative Models:
\begin{enumerate}
    \item DTAI has a large derivative with respect to an individual objective when it is currently underperforming the corresponding performance target. Conversely, the objective has an exponentially decaying derivative with respect to an individual objective as it further outperforms the objective's performance target. 
    \item DTAI is differentiable across the entire space of possible performance and constraint values and its derivative is continuous. This allows it to be used directly in the optimization functions of gradient-based generative methods. 
    \item DTAI is easy to calculate, with computational cost scaling linearly with the number of design objectives. 
    \item DTAI is bounded between zero and one. The gradient of DTAI is also bounded given a particular set of $\alpha$ and $\beta$ parameters. 
    \item The scoring function allows for easy weighting of objectives and modulation of score decay, which allow for easy and precise customization by the designer. 
\end{enumerate}
We note that this scoring metric is intended for use in targeted inverse design applications, where generating high-performing designs to meet a specific set of performance targets is the overarching goal. This score is poorly suited for quantification of performance space coverage or unconditional design synthesis. 

\subsection{Hypervolume (HV)} \label{sec:hypervolume}
Hypervolume is a useful metric for simultaneously quantifying performance space coverage and overall design optimality of a collection of design candidates. When calculating the hypervolume metric of a set of designs, we consider an T-dimensional space where T is the number of design objectives. The hypervolume is given by the volume of the union of all points within some hypercube spanning one of the designs and a common reference point. 

Hypervolume is a frequently used metric in multi-objective optimization and typically aims to quantify a solution set's proximity to the Pareto front~\cite{bringmann2013approximation, auger2009theory}. In the context of targeted inverse design (i.e. designing a product for a specific collection of design targets) hypervolume suffers from several limitations. In particular, hypervolume tends to 1) Over-reward further optimization of objectives that have already exceeded performance targets 2) Under-reward focused performance improvements to meet performance targets 3) Ignore the relative importance of different targets. 4) Require non-negligible computational expense. Though implanting design targets as the reference point for hypervolume calculation is a method to ensure that only designs that exceed all performance targets are scored, this method remains rather inflexible since designs that nearly miss performance targets are treated the same as designs that drastically miss them. For this reason, we select a reference point far below the performance targets of each objective, which we discuss further in Sec.~\ref{sec:methodology}.

\subsection{Design Diversity (DSD \& PSD)}
In generative design problems, we may seek to generate a diverse set of design candidates to give a designer a variety of design possibilities. Given a set of generated designs, we can score the diversity of each design by calculating its similarity to the other designs in the set and averaging these values. Mathematically, 
\begin{equation}
    s_{div,\,i}=\frac{1}{n-1}\sum_{j\in P} (\phi(y_i, y_j))
\end{equation}
Here, $s_{div,\,i}$ is the diversity score, $y$ is the set of generated designs, and $y_i$ refers to the $i^{th}$ design in the generated set. $n$ is the number of sampling iterations, $P$ is a randomly selected set of $n$ designs from $y$, and  $\phi$ is the kernel function calculating similarity, in this case, Euclidean Distance. Design diversity can be calculated in the \textit{design space} by calculating the similarity between parametric design vectors, which we call Design Space Diversity (DSD). Design diversity can also be quantified in the \textit{performance space} by calculating the similarity between vectors of design performance values, which we call Performance Space Diversity (PSD).

% Similarity metrics to quantify design diversity, we an approach based on similarity kernels borrowed from \cite{chen2021padgan} and \cite{chen2021mopadgan}. This diversity score computes an estimated aggregate diversity score by randomly sampling generated designs, computing a similarity matrix, computing the log determinant of this matrix, then averaging scores over multiple sampling iterations. Mathematically, 
% \begin{equation}
%     s_{div}=\frac{1}{n}\sum_{i-0}^n \log \det(L_{S_i})
% \end{equation}
% Here, $S_{div}$ is the diversity score, $n$ is the number of sampling iterations, and $S_i$ is the set of designs sampled at a particular iteration $i$. $L_{S_i}$ is the similarity matrix of $S_i$ computed using some kernel function, in this case Euclidean Distance. Design diversity can be calculated in the \textit{design space} by calculating distance between parametric design vectors, which we call Design Space Diversity (DSD). Design diversity can also be quantified in the performance space by calculating distance between vectors of design performance values, which we call Performance Space Diversity (PSD). 

\subsection{Design Novelty (DN)}
Quantifying a design's novelty can be important in real-world design tasks where intellectual property is an important concern. We adopt an approach based on~\cite{chen2021padgan}. Given a set of designs, we can score the novelty of each design by calculating its similarity to the designs in the original dataset and finding the minimum value (similarity to the most similar original design). Mathematically, 
\begin{equation}
    s_{DN,\,i}=\min_j(\phi(y_i, x_j))
\end{equation}
Here, $s_{DN,\,i}$ is the novelty score, $y_i$ refers to the $i^{th}$ design in the generated set, and $x_j$ refers to the $j^{th}$ design in the original dataset. Again, $\phi$ is the kernel function calculating similarity, for which we use Euclidean Distance. 

\subsection{Geometric Feasibility Rate (GFR)}
Generating physically feasible designs is an important consideration when evaluating generative methods. The Geometric Feasibility Rate is simply the ratio of total designs found to be feasible to the number of designs where feasibility status is known. By leveraging the simulation pipeline of the FRAMED dataset, we have a convenient way to explicitly quantify the Geometric Feasibility Rate of generated designs. Simply stated, feasible designs satisfy a set of predefined feasibility rules provided by FRAMED's authors and furthermore build into a valid 3D model provided. 

\subsection{Target Success Rate (TSR)}
Evaluating a generative method's ability to create designs that satisfy performance targets is critical. While the Design Target Achievement Index (DTAI) proposed is largely affected by a generated design's ability to satisfy performance targets, quantifying the raw fraction of performance targets satisfied is also a helpful reference metric. This ratio is expressed as the fraction of the design targets met or exceeded by any given design, weighted by the importance (as specified by the designer) of the targets. 

\begin{equation}
    s_{TSR, i}=\sum_k^T\alpha_kq_{i,k}
\end{equation}
\begin{equation}
    q_{i,k}=\begin{cases} 
          1 & p_{i,k}\geq t_k \\
          0 & p_{i,k}<t_k
    \end{cases}
\end{equation}
Here, T is the number of performance objectives. $\alpha_k$ is the importance of objective k and is the same parameter as the hyperparameter $\alpha$ in DTAI.

\subsection{Minimum Target Ratio (MTR)}
We may also want to evaluate the degree to which generated designs are meeting or failing performance targets. For any generated design, $y_i$ and objective $k$ consider the ratio $r_{i,k}$ between a designs performance $p_{i,k}$ and the performance target $t_{k}$. $s_{MTR, i}$ is defined as the minimum such ratio.
\begin{equation}
    s_{MTR, i}=\min_k\left(\frac{p_{i,k}}{t_{k}}\right)
\end{equation}
When the MTR is greater than one, it tells us by at least how much the performance in each objective outperforms the performance target. When the MTR is less than one, it tells us how far the design is underperforming the target in its most delinquent objective. Unlike the TSR, the MTR is not weighted by target importance.

\section{FRAMED Dataset} 
\begin{figure}
    \centering
    \includegraphics[width=0.5\textwidth]{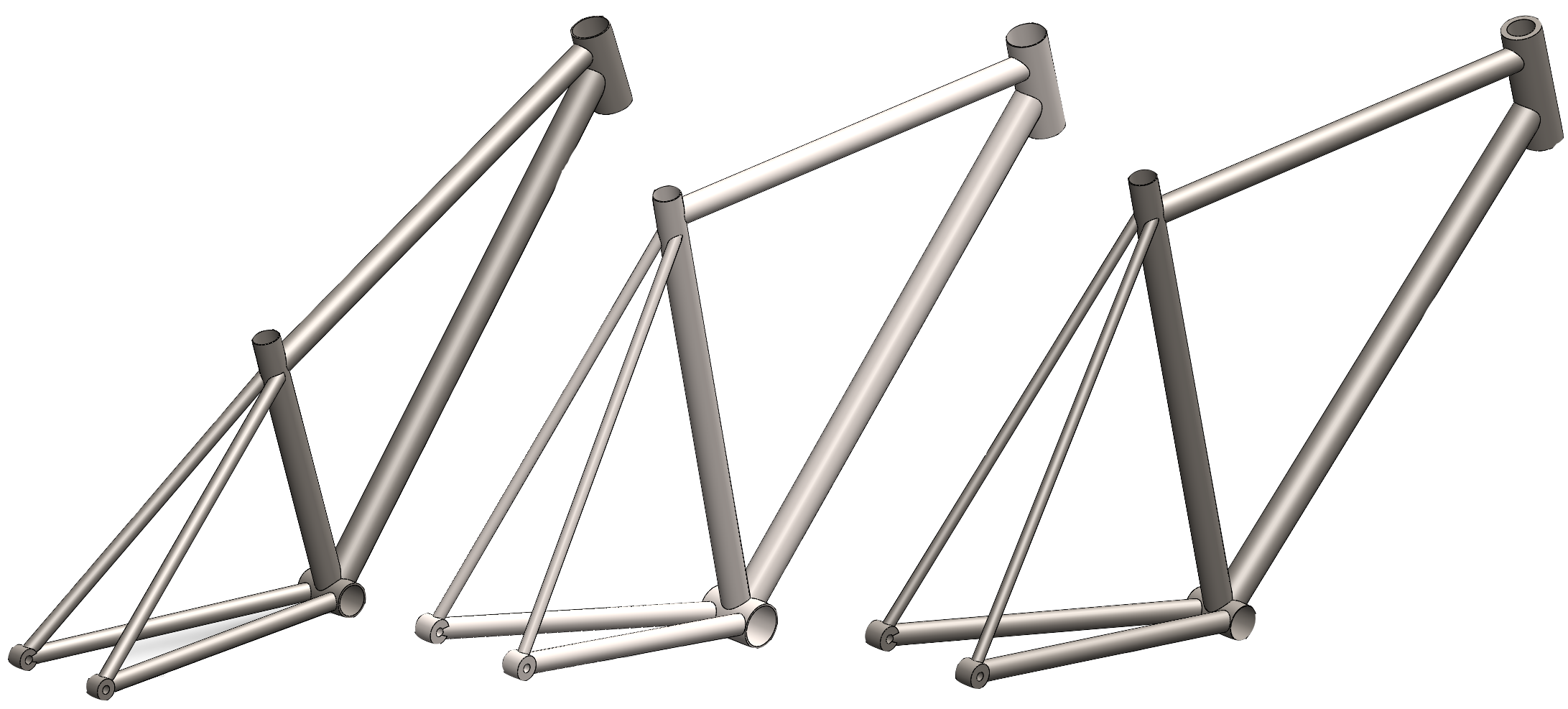}
    \caption{Examples of geometrically feasible bicycle frame models from the FRAMED dataset.}
    \label{fig:valid}
\end{figure}

For this study, we select the recently-introduced FRAMED dataset~\cite{regenwetter2022framed}, which consists of 4500 community-designed bicycle frame models parameterized over 37 design variables: tube lengths, diameters, and thicknesses, frame material, and frame junction locations. Three sample frame models from the dataset are rendered and shown in Figure \ref{fig:valid}. The FRAMED dataset provides model weight, and a set of 9 structural performance measures for a total of 10 objectives. These performance measures are derived from simulation results calculated in Finite Element Analysis (FEA) under three loading cases: 
\begin{itemize}
    \item In-plane loading: Bottom bracket vertical and horizontal displacement, Dropout vertical and horizontal displacement, safety factor
    \item Transverse loading: Bottom bracket lateral displacement
    \item Eccentric loading: Bottom bracket vertical displacement and rotation, safety factor
\end{itemize}

These performance metrics constitute a challenging mixed optimization problem with competing objectives. This bicycle frame optimization task is an active research field with significant commercial investment. 

\begin{figure}
    \centering
    \includegraphics[width=0.5\textwidth]{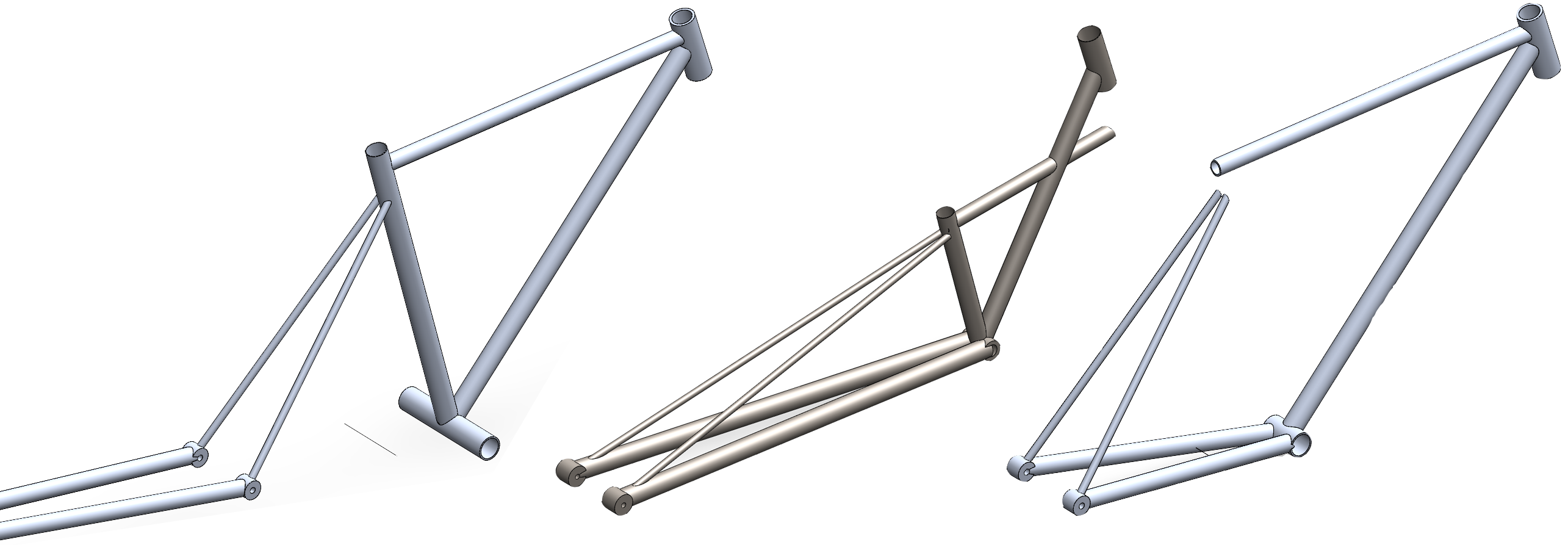}
    \caption{Examples of geometrically infeasible bicycle frame models from the FRAMED dataset.}
    \label{fig:invalid}
\end{figure}

\begin{figure}
    \centering
    \includegraphics[width=0.5\textwidth]{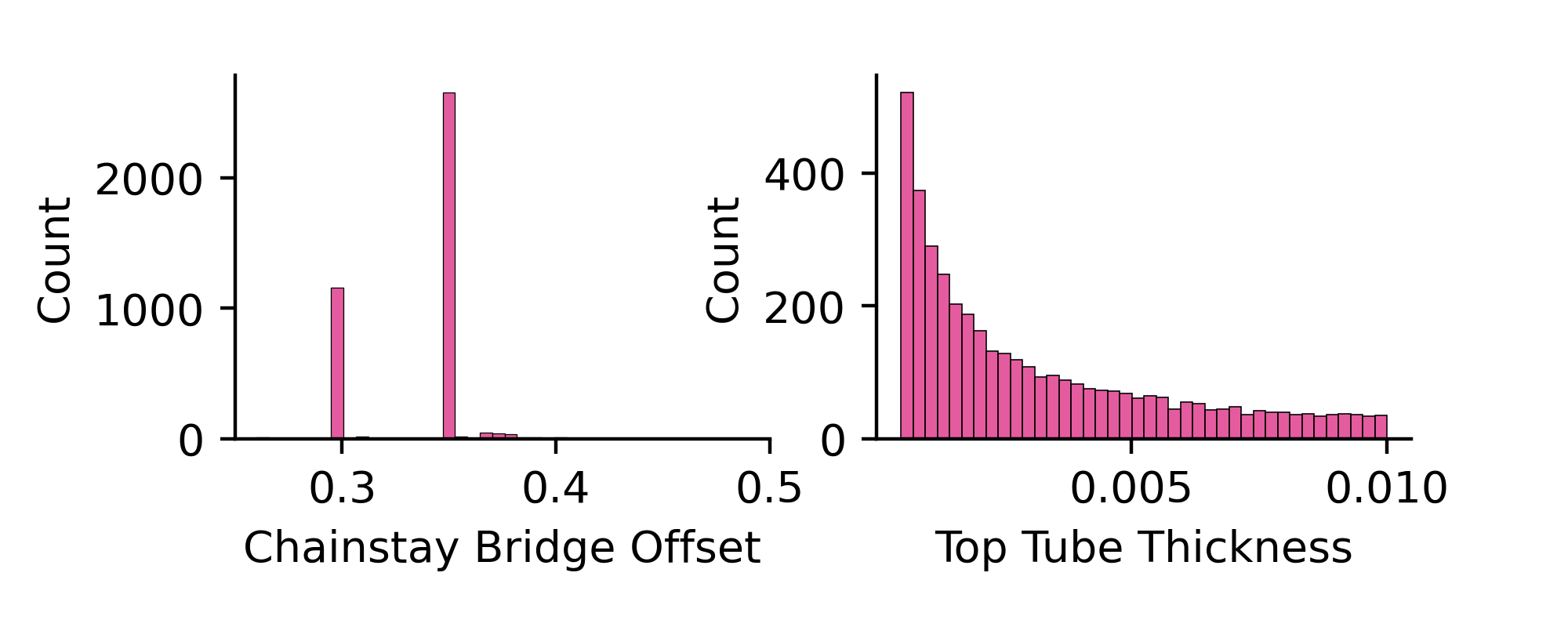}
    \caption{Examples of multimodal and skewed parameter distributions in FRAMED dataset.}
    \label{fig:distributions}
\end{figure}

Compared to other datasets frequently used to test and benchmark generative methods, such as the UIUC airfoil database, FRAMED's ten objectives make for a significantly more complex multi-objective optimization problem. Additionally, FRAMED considers the problem of design feasibility by sorting frames into feasible and infeasible designs. Frames are determined to be infeasible through a systematic check of geometric flaws. These flaws include negative tube lengths and thicknesses, triangles with one side longer than the sum of the counterparts, or dimensions that would cause parts not to connect. Some frames tha pass the explicit checks fail to generate into a proper 3D model when built in the FEA simulation software, the reasons for which are highly unpredictable and nearly impossible to exhaustively list. Three frames that build incorrectly in the FEA simulation software are shown in Figure \ref{fig:invalid}. While most of these frames are invalid, their status is uncertain, so they are discarded. FRAMED also presents a challenge for DGM training in that it features mixed datatypes (categorical and continuous), and features data with multimodal and skewed distributions, as seen in Figure~\ref{fig:distributions}. Unlike most DGM models used for image data, DGM models for tabular data are less well understood and are generally considered more challenging to train.

FRAMED also shares the simulation methodology, code, and 3D CAD model files, which gives us the capability to simulate models we generate while testing the various Deep Generative Models discussed in this work. Critically, this allows us to evaluate and benchmark these generative methods with a variety of evaluation metrics, most of which are based on performance values calculated using FRAMED's simulation methodology.

\section{Generative Methods}
In this section, we present the Deep Generative Models evaluated in this paper, discussing the core functionality and key innovations of each method. We additionally present two trivial baseline cases which we test alongside the Deep Generative Models. 

\subsection{Baselines} 
To establish baselines, we consider two trivial ``generative methods'' against which to benchmark more advanced frameworks. The first of these is a simple random sampling from the dataset. The second of these is a linear interpolation between dataset designs. Mathematically, we generate an interpolant, $d_{int}$ by randomly selecting two designs from the dataset, $x_1$ and $x_2$ as well as in interpolation factor, $\gamma \in [0,1]$. Our interpolant is given by: 
\begin{equation}
    x_{int}=\gamma\,x_1+(1-\gamma)x_2
\end{equation}
While both of these random sampling methods may be practical methods to select designs, we expect the Deep Generative Models tested to outperform these baselines in most metrics. 

\subsection{Generative Adversarial Network} \label{sec:GAN}
Introduced in 2014, the Generative Adversarial Network (GAN)~\cite{goodfellow2014generative} has become a staple across numerous fields, largely due to its unprecedented performance in generating convincing human faces~\cite{karras2020analyzing}. The GAN consists of a generator network and a discriminator network that play an adversarial game in which the discriminator attempts to distinguish real samples from the dataset from generated samples from the generator, while the generator attempts to fool the discriminator. Though many variations of the GAN have been proposed, we use a standard GAN as a baseline. 

\subsection{Tabular Generation Algorithms} \label{sec:tabgen}
The task of generating tabular parametric data frequently occurs in non-design-related applications. As such, many algorithms have been proposed to address the tabular generation problem, though, to our knowledge, all have focused on statistical similarity as the key training objective. We test two well-established methods, the Tabular Variational Autoencoder (TVAE) and the Conditional Tabular Generative Adversarial Network (CTGAN)~\cite{xu2019modeling} introduced by Xu~\etal~in 2019. CTGAN and TVAE established state-of-the-art performance on tabular data generation for their respective classes of generative methods (GAN and VAE). These methods serve primarily as a benchmark for high-performance methods that do not explicitly address performance-aware design generation.

While we refer the reader to Xu~\etal's paper~\cite{xu2019modeling} for details, we summarize the key innovations below. CTGAN and TVAE primarily address key challenges of tabular data, such as multimodal or skewed distributions, mixed datatypes (continuous and categorical), and skewed categories, all of which are common in the FRAMED data. CTGAN and TVAE both implement advancements to better handle both continuous and categorical datatypes in tabular data. 

To better learn continuous data, the authors implement a method that they call ``mode-specific normalization.'' This approach assumes that continuous variables fall into a Gaussian Mixture distribution and learns a Variational Gaussian Mixture Model. Sampled parameters are then probabilistically assigned to a particular mode and represented in terms of this assigned mode and the corresponding p-value within that mode's Gaussian distribution. This allows CTGAN and TVAE to more reliably learn complex distributions over continuous parameters.

To better learn categorical data, CTGAN trains conditionally using possible values of every categorical parameter in the data as the training condition. This prevents CTGAN from ignoring particular data categories, avoiding mode collapse. TVAE improves performance on categorical data simply by employing mixed activation functions in the final layer of its decoder. In particular, categorical variables are generated using a softmax activation. The combination of advancements for categorical and continuous parameters in tabular data makes CTGAN and TVAE particularly effective in mixed-datatype tabular generation problems.

In our testing, CTGAN and TVAE are trained on the original dataset designs without any one-hot encoding and do not utilize the performance data. We train for a maximum of 2000 epochs. 

\subsection{Performance-Augmented Diverse GAN (PaDGAN)} \label{sec:padgan}
Introduced in 2021 by Chen \& Ahmed, the Performance-Augmented Diverse GAN (PaDGAN)~\cite{chen2021padgan} specializes in performance- and diversity-aware design generation. PaDGAN demonstrated convincing synthesis performance on a variety of synthetic datasets as well as an airfoil design problem, generating a diverse set of samples that significantly exceeds the original dataset in average performance. 

We refer the reader to the original paper by Chen \& Ahmed for implementation details but summarize the key innovations below. PaDGAN implements an auxiliary training loss based on a Determinantal Point Process (DPP), which calculates a matrix over a batch of designs based on the similarity of designs in the batch and the quality (performance) values of the designs. The DPP loss is then calculated from this DPP matrix using a scaled log determinant. PaDGAN relies either upon a deterministic quality function or a differentiable approximation for the quality known as a surrogate model. 

An essential component of PaDGAN is a rapid performance evaluation method, which can be queried during training to evaluate generated design candidates. Our performance objectives have no deterministic relations based on design parameters and simulating designs in the training loop is too costly, so we fit a surrogate model to approximate the ten adjusted performance values and provide this model to PaDGAN to query during training. We select a deep neural network with four hidden layers of 200 neurons and batch-normalization and rectified linear unit activation functions after every hidden layer. This surrogate achieves a coefficient of determination of 0.681 on the entire dataset, implying that the regression fit is moderately strong. While FRAMED's authors present a higher-performing surrogate based on ensembles of individual regressors, including non-differentiable tree-based regressors, PaDGAN requires that the loss calculation be fully differentiable. Identifying higher-performing differentiable surrogates is an area for future work. 

PaDGAN inherently optimizes for a single objective, but Chen \& Ahmed extend PaDGAN to Multi-Objective PaDGAN (MO-PaDGAN) with a method for combining multiple objectives into a single aggregate score. MO-PaDGAN's proposes computing a randomly averaged weight of the different performance values for each sample propagated through the scoring function. The authors' primary motivation for this approach is the exploration of different regions of the design space's ``Pareto-front.'' The Pareto-front is a boundary of the design space consisting of all potentially ``optimal'' designs where any improvement in one objective must come at the expense of another. Since PaDGAN and MO-PaDGAN requires that objectives be maximized, we try taking a simple inverse to convert deflection and weight objectives into maximization problems. We find that MO-PaDGAN trains unstably since it requires that performance scores within a batch be of similar magnitude. This instability traces back to unbounded individual objective scores since objectives can be arbitrarily close to zero before inversion. To avoid this instability we propose an alternate approach which bounds resultant scores:

\begin{equation}
    s_{i,k}=\frac{p_{k, med}}{p_{k, med}+p_{i,k}}
\end{equation}
Here, $p_{k, med}$ is the median performance in objective $k$ of all dataset designs, while $p_{i,k}$ is the performance of a particular design, $i$ to be scored. Scores can then be weighted using the below function, where $w_k$ are random weights in a given range, say, $[0,1]$. 
\begin{equation}
    s_{MO,i}= \frac{\sum_k^Ts_{i,k}*w_k}{\sum_k^T w_k}
\end{equation}

\subsection{DTAI and Classfier-augmented PaDGAN}
While the proposed approach allows us to use MO-PaDGAN on FRAMED, we propose to replace $s_{MO}$ scores with Design Target Achievement Index (DTAI) scores instead. DTAI is better suited for targeted inverse design, spending less time exploring regions of the design space and focusing on direct optimization given a specified set of minimum performance targets. 

We further augment PaDGAN with an auxiliary classifier trained to classify feasible bicycle frames. We scale our DTAI score by this classifier's predicted likelihood of the given frame being geometrically valid and use this scaled score in PaDGAN's DPP loss. 

\begin{equation}
    s_{Tot,i}=s_{DTAI,i}*Q(y_i)
\end{equation}
Here, Q is a classifier that predicts the likelihood of a design, $y_i$ to be valid. Like the regressor, we are limited to differentiable surrogates and select a neural network that achieves a classification F1 score of 0.71. PaDGAN implements two tuning parameters, $\gamma_0$, which adjusts the weight of performance score in the loss function (compared to diversity), and $\gamma_1$, which adjusts the weight of the combined performance and diversity loss in the overall training respectively. We set $\gamma_0=5$, $\gamma_1=0.5$, and train for 50,000 iterations.

\section{Methodology} \label{sec:testing}
To test each of the methods, we first train the framework on only the valid designs in the full FRAMED dataset. We then generate 250 designs using the trained model, filter out initially infeasible designs, then simulate all remaining designs using the FEA simulation framework proposed in FRAMED. We then evaluate all feasible designs on the evaluation metrics proposed (save for Geometric Feasibility Rate, which is evaluated as the ratio of feasible to infeasible designs). We repeat this process three times for every method and report median scores over the three instantiations. 

\subsection{Data Preprocessing}
FRAMED's raw deflection values are given as absolute deflections, but we simplify to only consider deflection magnitudes. Since we assume the maximization of a set of positive objectives, we invert deflection and weight objectives (with the exception of PaDGAN when DTAI loss is not used -- See Section~\ref{sec:padgan}). We hereby refer to these modified performance values as the `adjusted performance values.' We discuss some of the training intricacies below.

\subsection{Evaluation Metrics}
Hypervolume calculations require a reference point. We select a reference point for hypervolume calculations such that each dimension's coordinate is equal to the FRAMED data's $1^{st}$ percentile objective score (worse than 99\% of FRAMED designs in each metric). Design Target Achievement Index, Target Success Rate, and Minimum Target Ratio all require a set of minimum performance targets. We select these targets to be equal to the $75^{th}$ percentile objective score (better than 75\% of FRAMED designs in each metric). Note that many objectives are difficult to simultaneously optimize since they inherently compete, so there is no guarantee that these minimum performance targets are even possible to simultaneously satisfy. This is a challenging objective, but realistic from an inverse design standpoint, as a designer may often set difficult or even impossible targets and expect a design that comes as close as possible. 

\section{Results}
\begin{figure*}[!ht]
    \centering
    \includegraphics[width=\textwidth]{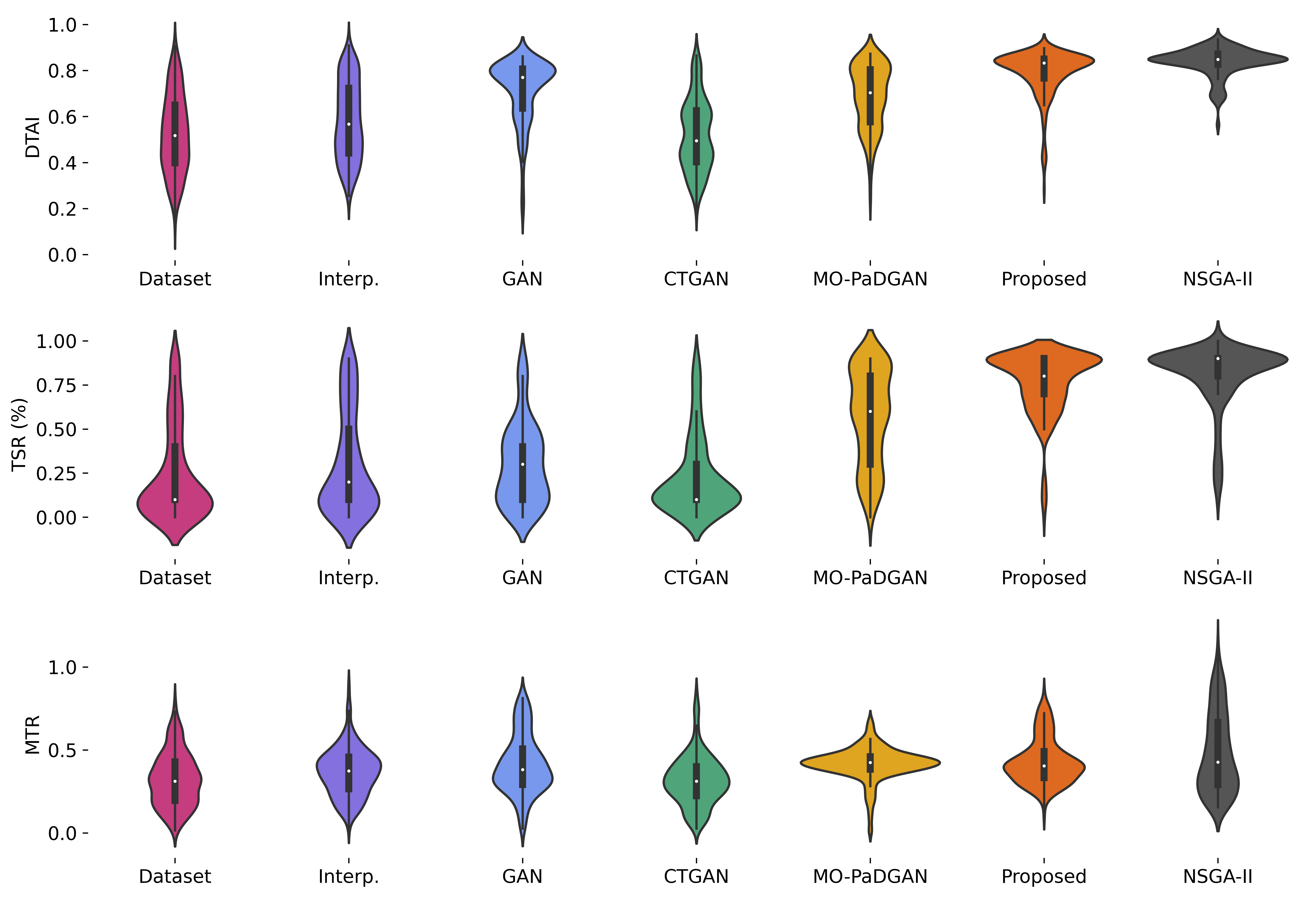}
    \caption{We demonstrate violin plots over the set of designs synthesized by each generative method for several metrics. Approximate distributions generated using Kernel Density Estimates are denoted with the colored curves. The black vertical boxes denote the interquartile range of the distribution, while the thin vertical line denotes the 5\%-95\% confidence interval. The median is denoted with a white point.}
    \label{fig:violins}
\end{figure*}

The overall results of our testing are shown in Table~\ref{tab:results}. We find that CTGAN and TVAE yield similar results, so we only present CTGAN results for simplicity. Table~\ref{tab:ablation} presents an ablation study analyzing the contributions of the DTAI loss and auxiliary classifier to the proposed method's performance. Figure ~\ref{fig:violins} presents Violin plots of the distribution of Design Target Achievement Index, Target Success Rate, and Minimum Target Ratio over the space of generated designs for baseline methods. A continuous distribution is approximated over the 250 designs generated by each method using a Kernel Density Estimate (KDE).

\begin{table*}[]
\centering
\caption{Deep Generative Models scored on the eight proposed evaluation metrics. Models from left to right: Randomly sampled design subsets from the FRAMED dataset (Dataset), Random Interpolation between FRAMED designs (Interpolation), Vanilla GAN (GAN) Conditional Tabular GAN (CTGAN), Proposed PaDGAN with DTAI and auxiliary Geometric Feasibility Classifier (Proposed)}
\label{tab:results}
\resizebox{0.9\textwidth}{!}{%
\begin{tabular}{|l|c|c|c|c|c|}
\hline
\textbf{Metric} & \textbf{Dataset} & \textbf{Interpolation} & \textbf{GAN} & \textbf{CTGAN} & \textbf{Proposed} \\ \hline
\textbf{Mean Design Space Diversity   (DSD)} & {\color[HTML]{000000} \textbf{8.60}} & {\color[HTML]{8DA500} \textbf{13.28}} & {\color[HTML]{A57B00} \textbf{10.08}} & {\color[HTML]{098D00} \textbf{14.12}} & {\color[HTML]{A53300} \textbf{9.48}} \\ \hline
\textbf{Mean Performance Space   Diversity (PSD)} & {\color[HTML]{000000} \textbf{3.80}} & {\color[HTML]{8DA500} \textbf{3.40}} & {\color[HTML]{A57B00} \textbf{2.71}} & {\color[HTML]{098D00} \textbf{3.95}} & {\color[HTML]{A57B00} \textbf{2.71}} \\ \hline
\textbf{Mean Design Novelty (DN)} & {\color[HTML]{000000} \textbf{0.00}} & {\color[HTML]{A53300} \textbf{7.17}} & {\color[HTML]{8DA500} \textbf{8.30}} & {\color[HTML]{A57B00} \textbf{8.02}} & {\color[HTML]{098D00} \textbf{9.96}} \\ \hline
\textbf{Geometric Feasibility Rate   (GFR) (\%)} & {\color[HTML]{000000} \textbf{100.0}} & {\color[HTML]{098D00} \textbf{100.0}} & {\color[HTML]{A53300} \textbf{65.2}} & {\color[HTML]{A57B00} \textbf{65.7}} & {\color[HTML]{8DA500} \textbf{95.9}} \\ \hline
\textbf{Hypervolume (HV) *10\textasciicircum{}-7} & {\color[HTML]{000000} \textbf{4.81}} & {\color[HTML]{098D00} \textbf{4.00}} & {\color[HTML]{A53300} \textbf{3.04}} & {\color[HTML]{A57B00} \textbf{3.54}} & {\color[HTML]{8DA500} \textbf{3.82}} \\ \hline
\textbf{Mean Design Target Achievement   Index (DTAI)} & {\color[HTML]{000000} \textbf{0.53}} & {\color[HTML]{A53300} \textbf{0.58}} & {\color[HTML]{8DA500} \textbf{0.71}} & {\color[HTML]{A57B00} \textbf{0.52}} & {\color[HTML]{098D00} \textbf{0.79}} \\ \hline
\textbf{Mean Target Success Rate (TSR)   (\%)} & {\color[HTML]{000000} \textbf{24.0}} & {\color[HTML]{8DA500} \textbf{31.3}} & {\color[HTML]{A57B00} \textbf{28.6}} & {\color[HTML]{A53300} \textbf{24.4}} & {\color[HTML]{098D00} \textbf{69.8}} \\ \hline
\textbf{Mean Minimum Target Ratio   (MTR)} & {\color[HTML]{000000} \textbf{0.32}} & {\color[HTML]{A57B00} \textbf{0.37}} & {\color[HTML]{8DA500} \textbf{0.41}} & {\color[HTML]{A53300} \textbf{0.32}} & {\color[HTML]{098D00} \textbf{0.43}} \\ \hline
\end{tabular}%
}
\end{table*}

\begin{table*}[]
\centering
\caption{Ablation Study Contrasting Proposed PaDGAN with DTAI training loss and auxiliary classifier against PaDGAN without DTAI (-DTAI), without the auxiliary classifier (-CLF), and MO-PaDGAN (-DTAI, -CLF). }
\label{tab:ablation}
\resizebox{0.8\textwidth}{!}{%
\begin{tabular}{|l|c|c|c|c|}
\hline
\textbf{} & \textbf{Proposed} & \textbf{-DTAI} & \textbf{-CLF} & \textbf{-DTAI, -CLF} \\ \hline
\textbf{Mean Design Space Diversity   (DSD)} & {\color[HTML]{098D00} \textbf{9.48}} & {\color[HTML]{A57B00} \textbf{6.21}} & {\color[HTML]{A53300} \textbf{5.80}} & {\color[HTML]{8DA500} \textbf{7.11}} \\ \hline
\textbf{Mean Performance Space   Diversity (PSD)} & {\color[HTML]{098D00} \textbf{2.71}} & {\color[HTML]{A53300} \textbf{2.24}} & {\color[HTML]{A57B00} \textbf{2.44}} & {\color[HTML]{8DA500} \textbf{2.62}} \\ \hline
\textbf{Mean Design Novelty (DN)} & {\color[HTML]{8DA500} \textbf{9.96}} & {\color[HTML]{A53300} \textbf{9.90}} & {\color[HTML]{098D00} \textbf{10.52}} & {\color[HTML]{A57B00} \textbf{9.94}} \\ \hline
\textbf{Geometric Feasibility Rate   (GFR) (\%)} & {\color[HTML]{098D00} \textbf{95.9}} & {\color[HTML]{A53300} \textbf{82.5}} & {\color[HTML]{A57B00} \textbf{83.2}} & {\color[HTML]{8DA500} \textbf{87.1}} \\ \hline
\textbf{Hypervolume (HV) *10\textasciicircum{}-7} & {\color[HTML]{098D00} \textbf{3.82}} & {\color[HTML]{098D00} \textbf{3.82}} & {\color[HTML]{A57B00} \textbf{3.40}} & {\color[HTML]{A53300} \textbf{2.95}} \\ \hline
\textbf{Mean Design Target Achievement   Index (DTAI)} & {\color[HTML]{8DA500} \textbf{0.79}} & {\color[HTML]{098D00} \textbf{0.80}} & {\color[HTML]{A57B00} \textbf{0.72}} & {\color[HTML]{A53300} \textbf{0.69}} \\ \hline
\textbf{Mean Target Success Rate (TSR)   (\%)} & {\color[HTML]{098D00} \textbf{69.8}} & {\color[HTML]{8DA500} \textbf{67.7}} & {\color[HTML]{A57B00} \textbf{51.4}} & {\color[HTML]{A53300} \textbf{48.4}} \\ \hline
\textbf{Mean Minimum Target Ratio   (MTR)} & {\color[HTML]{A57B00} \textbf{0.43}} & {\color[HTML]{098D00} \textbf{0.51}} & {\color[HTML]{8DA500} \textbf{0.46}} & {\color[HTML]{A53300} \textbf{0.42}} \\ \hline
\end{tabular}%
}
\end{table*}

\subsection{Feasibility Performance}
GAN and CTGAN perform poorly in generating feasible designs. Many of the infeasible designs generated result from negative tube thicknesses, explaining interpolation's strong performance in feasibility, since interpolating between positive thicknesses can never be negative. The difficulty to respect these feasibility constraints reflects the GAN and CTGAN's difficulty to learn sharp thresholds. In contrast the auxiliary classier in the proposed method provides the model with a strong gradient signal to avoid infeasible regions and shift the model's learned thresholds. 

\subsection{Target Achievement Performance}
Interpolation, GAN, and CTGAN all perform poorly in Target Success Rate (TSR), each achieving less than 1 in 3 design targets on average per design. In the proposed method, DTAI guides PaDGAN training to specifically achieve design targets, scoring just under 70\% design feasibility on average. It's distributions across designs are also significantly more consistent than competing objectives, as seen in Figure~\ref{fig:violins}. Using DTAI as a loss unsurprisingly improves DTAI scores, which themselves are highly reflective of target achievement performance. Interestingly, the standard GAN achieves moderately higher DTAI performance than the dataset, perhaps because it fails to capture less-conventional portions of the design space that may tend towards lower performance. PaDGAN with DTAI also achieves the best average Minimum Target Ratio (MTR), indicating that it is coming closer to targets across all objectives. This improvement in MTR is less pronounced than TSR or DTAI, potentially because MTR doesn't reflect the designer's weighting of objective importance, while the DTAI loss guiding training does. Nonetheless, the improvement of MTR over baselines is still significant. PaDGAN with DTAI even improves in Hypervolume scores over GAN and CTGAN. This indicates that despite the emphasis on targeted improvements to objectives, overall performance of the generates set across all objectives is increased. 

\subsection{Ablation Study} 
The ablation study shown in Table~\ref{tab:ablation} contrasts the proposed PaDGAN with DTAI and auxiliary classifier against variants without the classifier (-CLF) and using MO-PaDGAN's random objective weighting instead of DTAI (-DTAI). We note that the standard MO-PaDGAN (-DTAI, -CLF) was arguably the previous state-of-the-art in performance-aware design generation using DGMs. We outperform Mo-PaDGAN in every metric tested, with significant improvements in feasibility (95.9\% from 87.1\%), DTAI (0.79 from 0.69), Target Success Rate (TSR) (69.8\% from 48.4\%), and Hypervolume ($3.82E-7$ from $2.95E-7$). 

\subsection{Comparison to Multi-Objective Optimization}
While DGMs and multi-objective optimization frameworks are drastically different approaches to design generation, we briefly contrast our proposed method to Non-Dominated Sorting Genetic Algorithm II (NSGA-II)~\cite{deb2002fast}, a well-known multi-objective optimization framework. Due to the relatively small parameter space (37) and manageable objective space (10), applying current multi-objective optimization approaches like NSGA-II is viable on this dataset, though we note that such approaches may not work on datasets with larger parameter spaces and more objectives (whereas deep learning-based methods have been tried and proven on massive parameter spaces). Since evaluating actual performance values is too costly, even for an optimization-based approach, we allow NSGA-II to query performance predictions from our surrogate regression model. We further constrain NSGA-II to parameter ranges between the 2nd and 98th percentile of dataset values. This is necessary since: 1) Trivial solutions can be attained by minimizing the size of a bike frame. 2) Our surrogate model has limited generalization capability outside of the training data distribution, which would allow NSGA-II to exploit inaccurate predictions too easily. We train NSGA for 200 generations with a population of 250 and take its final population as `generated' samples. NSGA-II achieves a DTAI score of 0.84, TSR of 76.0, and MTR of 0.48, marking minor improvements of 0.05, 6.2, and 0.05 over our proposed PaDGAN with DTAI and auxiliary Classifier, which are reflected in the distributions shown in Figure~\ref{fig:violins}. Since NSGA-II attempts to find a diverse Pareto-Front, it also significantly improves in performance space diversity, with a PSD of 6.35. 

Despite the fact that optimization methods outperform DGMs in this particular bike frame design problem, improving DGM performance still has important ramifications since DGMs have numerous advantages over optimization-based approaches. For example, DGMs are broadly applicable to a larger and more complex design representations, particularly those using non-parametric data. Furthermore, DGMs may have the capability to scale to larger numbers of objectives, since the computational scaling of DTAI evaluation with objective count is linear. Practical testing of DTAI's scaling is a valuable research inquiry, however. 

\section{Limitations and Future Work}
We demonstrated the sweeping improvements that can be achieved by incorporating feasibility and performance into DGM training losses. As discussed in Section~\ref{Quality}, an inherent limitation with this approach is that the entire loss calculation procedure must be differentiable, which necessitates differentiable evaluation functions and surrogate models. Developing higher performing surrogates or even rapid differentiable numerical simulations is a promising approach to improve DGM performance. 

Since scoring methods requires time-intensive numerical simulation, we were limited in the tuning of the generative methods tested, and acknowledge that results may vary depending on the initialization of methods. While we mitigated this uncertainty by simulating three batches of 250 designs from three instantiations of each method and taking median values, moderate variability in results can be expected. Future work may include testing larger number of samples and more runs to improve the confidence in the results, as well as exploring effects of hyperparameter selections. 

This work leaves many avenues for further expansion. Training using only a high-performing subset of bicycle frames from the dataset may improve the performance of the DGMs tested, particularly the vanilla GAN and the tabular generation algorithms, as they factor in no notion of design performance. Testing the DTAI metric with other performance-aware generative frameworks besides PaDGAN would also be valuable. Finally, testing more methods and datasets would yield a more complete perspective. We encourage researchers to test new generative methods using our evaluation metrics on the FRAMED dataset or other datasets of their choice. 
\section{Conclusion}

We introduce a novel differentiable scoring metric called Design Target Achievement Index (DTAI) which allows Deep Generative Models to prioritize, meet, and exceed multi-objective performance targets. We augment a Performance-Augmented Diverse GAN with our DTAI objective and demonstrate significantly improved performance in design generation. We then further augment this PaDGAN with an auxiliary classifier to encourage the generation of feasible results. To benchmark our method, we evaluate a variety of Deep Generative Models, including the Multi-Objective PaDGAN, and specialized tabular generation algorithms CTGAN and TVAE. Methods are tested on a challenging bicycle frame design problem with 10 performance objectives. To rigorously evaluate methods for diversity, novelty, constraint satisfaction, overall performance, and feasibility, we propose a comprehensive set of evaluation metrics and score all tested methods on these metrics. The proposed PaDGAN with DTAI loss and auxiliary classifier significantly outperforms baselines in most performance objectives and further outperforms other PaDGAN variants in ablation studies. All in all, this work establishes a novel Deep Generative Framework that actively optimizes performance, diversity, feasibility, and target satisfaction to establish a new state-of-the-art in design generation using Deep Generative Models. 

\section{Acknowledgments}
We would like to thank Amin Heyrani Nobari for creating the Tensorflow 2.x version of PaDGAN which we modified to generate our results. We also acknowledge MathWorks for supporting this research. 

\bibliographystyle{asmems4}
\bibliography{bibliography}

\newpage

\end{document}